\pdfoutput=1

\documentclass[11pt]{article}

\usepackage[preprint]{coling}

\usepackage{colortbl}
\usepackage{times}
\usepackage{latexsym}
\usepackage{graphicx}
\usepackage{multicol}
\usepackage{cuted}
\usepackage{makecell}
\usepackage{diagbox}
\usepackage{booktabs}
\usepackage{mathptmx}

\usepackage{amsmath}
\usepackage{amssymb}

\usepackage[T1]{fontenc}

\usepackage[utf8]{inputenc}

\usepackage{microtype}

\title{\emph{Idea23D}: Collaborative LMM Agents Enable 3D Model Generation from Interleaved Multimodal Inputs}

\author{
\textbf{Junhao Chen\textsuperscript{1,2}\textsuperscript{*}},
\textbf{Xiang Li\textsuperscript{3}\textsuperscript{*}},
\textbf{Xiaojun Ye\textsuperscript{4}},
\textbf{Chao Li\textsuperscript{5}},
\textbf{Zhaoxin Fan\textsuperscript{6}\textsuperscript{\textdagger}},
\textbf{Hao Zhao\textsuperscript{1}\textsuperscript{\textdagger}}
\\
\textsuperscript{1}Institute for AI Industry Research (AIR), Tsinghua University, \\
\textsuperscript{2}Tsinghua Shenzhen International Graduate School, Tsinghua University,  \\
\textsuperscript{3}School of Software and Microelectronics, Peking University, \\
\textsuperscript{4}College of Computer Science, Zhejiang University, \\
\textsuperscript{5}College of Computer Science and Technology, Harbin Engineering University, \\
\textsuperscript{6}{Beijing Advanced Innovation Center for Future Blockchain and Privacy Computing,} \\
{School of Artificial Intelligence, Beihang University}
\\
}

\begin{document}

\maketitle

\renewcommand{\thefootnote}{} 
\footnotetext{
    \textsuperscript{*} Indicates Equal Contribution. \textsuperscript{\textdagger} Indicates Corresponding Author, email to \href{mailto:zhaohao@air.tsinghua.edu.cn}{zhaohao@air.tsinghua.edu.cn} and \href{mailto:zhaoxinf@buaa.edu.cn}{zhaoxinf@buaa.edu.cn}
}
\renewcommand{\thefootnote}{\arabic{footnote}}

\noindent\parbox{\textwidth}{\centering \textbf{Project Page: \url{https://idea23d.github.io/}}  \vspace{-0.9cm} }

\begin{strip}
    \centering
    \includegraphics[width=0.85\textwidth]{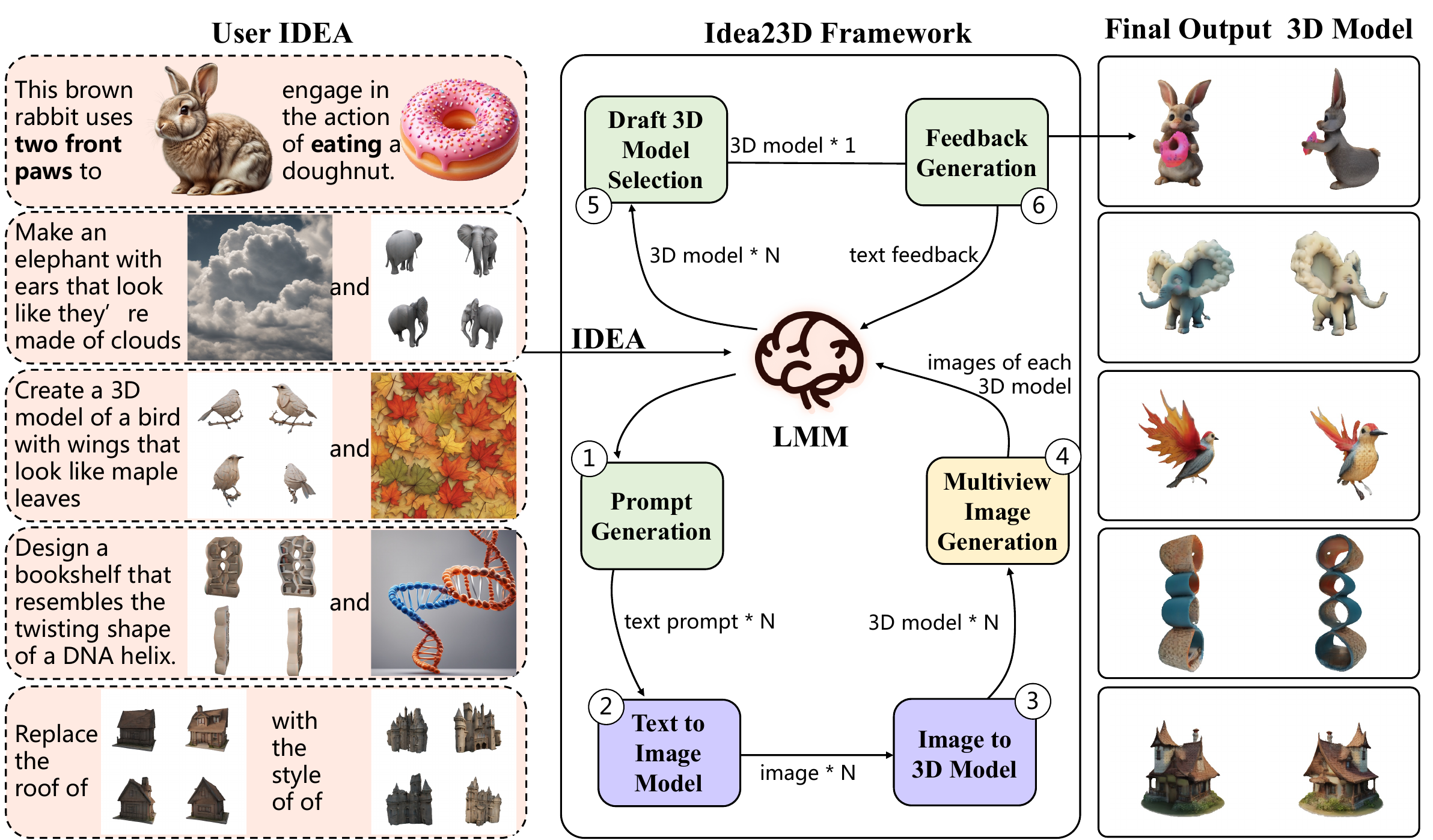}
    \captionof{figure}{
    The \emph{Idea23D} framework synergizes the capabilities of the Large Multimodal Model (LMM), Text-to-Image (T-2-I), and Image-to-3D (I-2-3D) models to transform complex multimodal input IDEAs into tangible 3D models. This process begins with the user articulating high-level 3D design requirements (\textbf{IDEA}). Following this, the LMM generates textual prompts (\textbf{Prompt Generation}) that are then converted into 3D models. These models are evaluated through a \textbf{Multiview Image Generation and Evaluation} process, leading to the \textbf{Selection of an Optimal 3D Model}. Subsequently, the T-2-I prompt is refined (\textbf{Feedback Generation}) using insights from the LMM. Additionally, an integrated memory module (see Sec.~\ref{sec_Memory_Module}), meticulously records each iteration, facilitating a multimodal, iterative self-refinement cycle within the framework. Note that this procedure is fully automatic \textbf{without any human intervention}.
    }
    \label{fig:example}
\end{strip}

\begin{abstract}

With the success of 2D diffusion models, 2D AIGC content has already transformed our lives. Recently, this success has been extended to 3D AIGC, with state-of-the-art methods generating textured 3D models from single images or text. However, we argue that current 3D AIGC methods still don’t fully unleash human creativity. We often imagine 3D content made from multimodal inputs, such as what it would look like if my pet bunny were eating a doughnut on the table. In this paper, we explore a novel 3D AIGC approach: generating 3D content from IDEAs. An IDEA is a multimodal input composed of text, image, and 3D models. To our knowledge, this challenging and exciting 3D AIGC setting has not been studied before. We propose the new framework Idea23D, which combines three agents based on large multimodal models (LMMs) and existing algorithmic tools. These three LMM-based agents are tasked with prompt generation, model selection, and feedback reflection. They collaborate and critique each other in a fully automated loop, without human intervention. The framework then generates a text prompt to create 3D models that align closely with the input IDEAs. We demonstrate impressive 3D AIGC results that surpass previous methods. To comprehensively assess the 3D AIGC capabilities of Idea23D, we introduce the Eval3DAIGC-198 dataset, containing 198 multimodal inputs for 3D generation tasks. This dataset evaluates the alignment between generated 3D content and input IDEAs. Our user study and quantitative results show that Idea23D significantly improves the success rate and accuracy of 3D generation, with excellent compatibility across various LMM, Text-to-Image, and Image-to-3D models. 
Code and dataset are available at \url{https://idea23d.github.io/}.

\end{abstract}

\section{Introduction}
\label{sec:intro}
Recently the success of 2D AIGC foundation models~\cite{rombach2022high,podell2023sdxl,zhang2023adding,shi2023mvdream,liu2023syncdreamer,openai2023dalle2,openai2023dalle3,yang2023idea2img,chen2023soulstyler,blackforestlabs2024flux1tools} has been translated to the 3D domain~\cite{liu2023zero, long2023wonder3d, poole2022dreamfusion,lin2023magic3d,voleti2024sv3d,xu2024instantmesh,tochilkin2024triposr,yang2024tencenthunyuan3d}. However, state-of-the-art models take RGB images or text prompts as inputs, which still fails to match the (wildest) creativity of humanity. Arguably, the nature of creativity is connecting (seemingly unrelated) dots that share intrinsic harmony. So we propose a novel 3D AIGC setting in which all prior arts fail: generating textured 3D models from \textbf{IDEAs}. The formal definition of an IDEA is an interleaved sequence of multi-modal inputs, covering modalities like text, images and 3D models. 
We show some typical IDEAs in Fig.~\ref{fig:example}, which contain a text prompt and images or a 3D model. 
We use rendered images to represent the 3D model in Fig.~\ref{fig:example}.
This kind of IDEAs come into our minds now and then in the daily life: Someone takes a doughnut into the room and looks at her pet rabbit, imaging what it may look like if the rabbit is eating the doughnut using front paws. This is the moment that creativity happens but as far as we know, no existing 3D AIGC foundation models can take this kind of IDEAs as input. More different types of IDEAs are shown in Fig.~\ref{fig:sample_demos} and Fig.~\ref{fig:morecase}.

Existing methods in text-based 3D model generation~\cite{wang2023prolificdreamer, poole2022dreamfusion, lin2023magic3d, liu2023zero, qian2023magic123, yang2024tencenthunyuan3d}, known as Text-to-3D (T-2-3D), have made progress in certain aspects such as fidelity, but they still face substantial challenges, particularly when dealing with complex and abstract interleaved multimodal inputs (IDEAs). 
A potential solution for adapting existing T-2-3D methods to handle IDEA inputs is converting the images and 3D models in the IDEAs into natural language descriptions. However, this approach is time-consuming and requires a certain level of expertise from the user.

Our proposal is to use a multi-agent collaboration framework. LLM (Large Language Model) agent systems~\cite{xi2023rise, AgentGPT2023, xagent2023, richards2023auto, gong2023mindagent, liu2023dynamic, gu2022don, liu2023llm, chen2023chatcot, gou2023critic} have already demonstrated remarkable effectiveness in solving complex natural language processing tasks, suggesting their potential application in the T-2-3D domain. There are already some recent successful methods that leverage LLM agents for computer vision applications~\cite{gupta2023visual, suris2023vipergpt,wei2024editable, yang2023llm, huang2023embodied, yang2023idea2img, mmad2024}. They exploit the generic methodology of prompting LLM agents to write codes and invoke existing computer vision functions. We inherit this methodology but exploit LMMs (Large Multimodal Models) as agents because the visual inputs are critical in understanding IDEAs.

However, designing a LMM agent system to generate 3D models from IDEAs is not straightforward and presents its own set of challenges, especially effective integration and understanding of multimodal inputs. To tackle these challenges, we propose \emph{Idea23D}, a framework that employs three different agents based upon the powerful LMM, for iterative self-improvement in automated 3D design and generation. Specifically, as shown in Fig.~\ref{fig:example}, \emph{Idea23D} consists of three LMM agents (green boxes indexed by 1,5,6) acting in the roles of prompt generation, model selection and feedback reflection. 

\emph{Idea23D} combines the capabilities of LMM agents and other multimodal algorithmic modules (purple boxes indexed by 2,3 and yellow box indexed by 4 in Fig.~\ref{fig:example}) to generate textual prompts from interleaved user inputs (IDEAs), which are then converted into 3D models. This process involves iterative refinement, utilizing a memory module to record each iteration and support continuous improvement. As shown in Fig.~\ref{fig:sample_demos} and Fig.~\ref{fig:morecase}, \emph{Idea23D} can generate high-quality 3D models that well align input \textbf{IDEAs in a fully automatic manner} while caption-based baselines constructed from prior T-2-3D models can hardly generate meaningful results. 

Our qualitative comparisons and quantitative experiments demonstrate the effectiveness of \emph{Idea23D}, especially in handling complex and challenging IDEA inputs.
The contributions of this paper are as follows.

\renewcommand{\labelenumi}{(\theenumi)}
\begin{enumerate}
    \item \emph{\textbf{Idea23D}} is the first work to achieve the transformation of high-level, abstract user IDEAs (multimodal interleaved inputs) into concrete 3D models, realizing a fully automated 3D AIGC task.
    \item Surpassing the capabilities of existing LLM agent systems in 3D AIGC, \emph{\textbf{Idea23D}} demonstrates the effectiveness of LMM-based agents in improving, evaluating, and validating multimodal content for 3D model generation.
    \item Proposes a challenging evaluation dataset \textbf{Eval3DAIGC-198} with multimodal inputs, and proves the effectiveness of \emph{Idea23D} through comprehensive user preference studies and 3D visual caption experiments.
\end{enumerate}

\section{Related Works}
\label{append:Related_Works}

\subsection{Self-refining Agents} 
Our research builds on the self-refinement capability of large language models (LLMs). Recent studies show that LLMs are effective at self-refinement, when exploring unknown environments and tasks~\cite{xi2023rise, madaan2024self, pan2023automatically, shinn2023reflexion, lee2023platypus}. 
For example, projects such as \emph{Self-refine}\cite{madaan2024self} and \emph{Reflexion}\cite{shinn2023reflexion} utilize LLMs to iteratively critique their outputs and use feedback to improve predictions, resulting in significant performance improvements in natural language processing tasks.
However, these approaches mainly excel in tasks dealing with natural language descriptions~\cite{shridhar2020alfworld}. In contrast, our \emph{Idea23D} project employs an iterative self-refinement system based on LMM in a multimodal environment, especially for interleaved inputs of text, images, and 3D models (IDEAs), in a different way other than the traditional approach focusing solely on natural language inputs.

\subsection{Large Multimodal Model} 
Building on the Large Language Model (LLM), the development of the Large Multimodal Model (LMM) marks an important evolution from unimodal to multimodal processing capabilities. Initial LLMs, such as the GPT family, focused on the generation of textual data, demonstrating superior capabilities in understanding and creating natural language~\cite{brown2020language, radford2019language, openai2023gpt4, bai2023qwen1, bai2023qwen, zhang2023video, he2024ma}.
As the field evolves, the processing capabilities of the models expanded from pure text to include multimodal data including images, audio, and video~\cite{ramesh2021zero, radford2021learning}.
For example, CLIP~\cite{radford2021learning} was the first to achieve cross-modal alignment between images and text, enabling cross-modal understanding between text and image. Some models~\cite{liu2023visual, xu2024llavacot, chen2024expanding-InternVL25, Qwen-VL, Qwen2-VL}, demonstrates multimodal understanding and generation of mixed text, image, and video inputs.
Further, projects such as \emph{Uni-3D}~\cite{zhang2023uni} and \emph{SDFusion}~\cite{cheng2023sdfusion} extend this concept to the 3D design domain, enabling good 3D model understanding, generation and reconstruction. LMM has also enabled open-set scene understanding in various settings \cite{tian2023unsupervised, li2022toist, li2023understanding, liu2023delving, jin2023adapt}. 
Despite the progress made in the field of 3D understanding and generation, current LMMs can still only handle input content at the text and image level, and struggle to handle multimodal high-level inputs containing text, images, and 3D models (IDEAs in our case). In contrast, the \emph{Idea23D} project takes a much larger step forward in multimodal 3D model generation. Our system is capable of processing not only single-modal inputs, but also composite multimodal inputs containing text, images, and 3D models at the same time.

\subsection{Extensions of T-2-3D Models}
There is already a large literature extending seminal T-2-3D models, including variants enabling T-2-3D models to better follow user prompts~\cite{black2023training, chefer2023attend, feng2022training} , refine keywords in T-2-3D prompts for better visual quality~\cite{gu2023systematic}, support for additional image inputs for image processing~\cite{brooks2023instructpix2pix, kawar2023imagic}, support for additional 3D model inputs for 3D model processing~\cite{cheng2023sdfusion}, 3D style migration~\cite{pan2023efficient, ma2014analogy, segu20203dsnet}, visual concept customization~\cite{wei2023elite, kumari2023multi}, using image generation models to generate 3D textures~\cite{chen2023text2tex, zeng2024paint3d, chen2024ultraman, yu2024texgen}, and more. 
Going even further, \emph{Idea23D} offers users a more natural way to design and create the 3D content they want. Similar to \emph{Visual ChatGPT}~\cite{wu2023visual}, which extends \emph{ChatGPT}'s~\cite{oepnai-gpt-turbo} ability to understand and generate 2D images, \emph{Idea23D} is designed to provide a more unified and broadly applicable framework for automated 3D model design and generation. 
\emph{Idea23D} extends the multimodal input and 3D model understanding and generation capabilities of the LMM~\cite{openai2023gpt4} using the T-2-I model~\cite{shi2020improving} and I-2-3D model~\cite{ramesh2021zero}, as well as the multimodal input capabilities of the T-2-3D model~\cite{wang2023prolificdreamer, poole2022dreamfusion, lin2023magic3d, wang2023score, metzer2023latent, tsalicoglou2023textmesh, qian2023magic123, liu2023zero,long2023wonder3d} and other standard off-the-shelf algorithmic modules like multiview rendering.

\section{\emph{Idea23D} Framework}
\begin{figure*}[htbp]
\vspace{-1cm}
  \centering
    \includegraphics[width=0.9\linewidth]{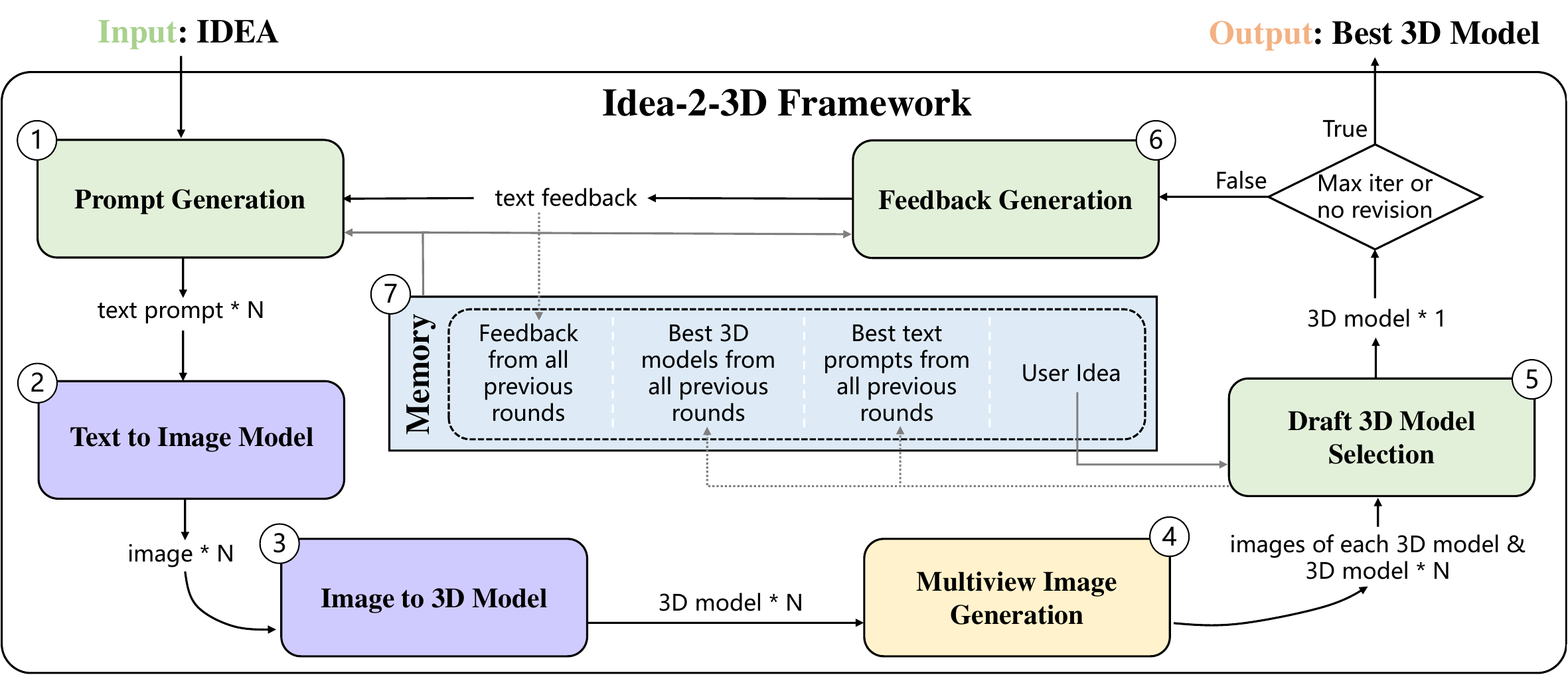}
    \caption{Overview of the framework of \emph{Idea23D}, which employs LMM agents to unleash the T-2-3D model's potential through iterative self-refinement to provide better T-2-3D prompts for the input user IDEA. Green rounded rectangles indicate steps completed by LMM agents. Purple rounded rectangles indicate T-2-3D modules, including T-2-I models and I-2-3D models. The yellow rounded rectangle indicates the off-the-shelf 3D model multi-view generation algorithm. The blue color indicates the memory module, which saves all the feedback from previous rounds, the best 3D model, and the best text prompt. Note that this cycle is \textbf{fully automatically} executed by LMM agents, without any human intervention.}
    \label{fig:overview_famework}
\end{figure*}

The \emph{Idea23D} framework represents a novel approach to generate detailed 3D models from high-level, abstract multimodal inputs (IDEAs), shown in Fig.~\ref{fig:overview_famework}. It integrates three LMM-based agents and several off-the-shelf tools for agents to invoke. Specifically, three agents are responsible for prompt generation (green box indexed by 1 in Fig.~\ref{fig:overview_famework}), model selection (green box indexed by 5 in Fig.~\ref{fig:overview_famework}) and feedback generation (green box indexed by 6 in Fig.~\ref{fig:overview_famework}). Two foundation models for T-2-I (purple box indexed by 2 in Fig.~\ref{fig:overview_famework}) and I-2-3D (purple box indexed by 3 in Fig.~\ref{fig:overview_famework}) are exploited together to turn natural language prompts into textured 3D models. As shown in Tab.~\ref{tab:result_tab}, various foundation model variants are evaluated for comprehensiveness. A unique memory module enhances the system, retaining insights from previous iterations to optimize future outputs.

The process begins with the LMM (\textbf{agent 1}) converting multimodal IDEAs into T-2-I prompts, which facilitate the creation of preliminary 3D drafts. These drafts undergo a selection process (\textbf{agent 5}) where the best 3D model is either considered finalized or subjected to further refinement based on LMM feedback (\textbf{agent 6}). The cycle continues until the model satisfies the user's IDEA (as judged by \textbf{agent 5}) or reaches a pre-set iteration limit. This framework, illustrated in our system diagram (Fig.~\ref{fig:overview_famework}), enables continuous improvement through a loop with automatically generated feedbacks.

\subsection{Multimodal IDEA Input}
The user-provided IDEA $X$ encapsulates the overarching 3D modeling requirement, represented by a multimodal input set composed of text, images, and 3D models.
$X = \{T, I, M\}$, where $T$ is a suite of textual directives encompassing descriptive phrases, keywords, and design specifications that \textbf{may refer to both} 2D and 3D information. $I$ is an assortment of images such as reference shots, diagrams, or related illustrations. $M$ is a compilation of 3D models, including pre-existing constructions or particular design ingredients furnished by the user.
Each element of $X$ reflects a facet of the user's design intent, and their aggregation forms a comprehensive IDEA. This input specification is designed to capture the user's intent in a multi-faceted and multimodal manner, laying the ground for the subsequent procedural stages, including initial prompt formulation and 3D model synthesis. As mentioned before, designing from this kind of abstract IDEAs is of great needs and our method is the first to fulfill this need.

\subsection{Initial Prompt Generation}
\label{sec_ipg}
Recall that, technically, the \emph{Idea23D} framework is tailored to convert complex multimodal user inputs $X$ into textual prompts for 3D model generation. Specifically, it employs the Large Multimodal Model (LMM) to comprehend and articulate these inputs into a format digestible by the Text-to-3D (T-2-3D) model. Addressing the high dimensionality of 3D models, \emph{Idea23D} leverages LMM's text and image processing strengths by representing 3D models $M$ as multi-view image sets $I'$ via a conversion function $\rm CM2I(*)$:

\begin{equation}I'={\rm CM2I}(M)\end{equation}

Specifically, Function $\rm CM2I(*)$ render each 3D model into six images, depicting the model from various perspectives: front, back, left, right, top, and bottom.
The aggregation of these images $I'$ with the original image set $I$ and textual components yields an augmented IDEA $X'$:

\begin{equation}X'=\{T,I'\cup I\}\end{equation}

The \textbf{LMM agent 1 for prompt generation} receives the IDEA $X'$ and generates $N$ specific descriptive instructions $\{P_0,P_1,... ,P_{N-1}\}$:

\begin{equation}P=f_{\rm LMM}(X',p_{\rm gen})\label{lmm1}\end{equation}

Here, $p_{\rm gen}$ represents a prompt facilitating the generation of T-2-3D prompts\footnote{Note here $p_{\rm gen}$ and P are prompts for LMMs and T-2-3D models, respectively. 
All prompts are in our code.}. This enables \emph{Idea23D} to extract and interpret not only direct descriptions (e.g. $X'$ with only $T$ is also acceptable) but also high-level concepts and intermixed modalities within the IDEA, such as images and 3D models.

In subsequent iterations as denoted by $\rm iter$, each T-2-3D prompt in the set of $P^{\rm iter}$ is used as input to the T-2-3D model to produce draft 3D models $D_i^{\rm iter}=T23D(P_i^{\rm iter})$, iteratively refining until the output aligns with the user's intents.

\subsection{3D Model Generation}
\label{sec_3D_Generation}
\emph{Idea23D} transforms the set of N text prompts $\{P_0, P_1, \ldots, P_{N-1}\}$ into an equivalent set of N 3D models $\{D_0, D_1, \ldots, D_{N-1}\}$ utilizing T-2-3D models. As outlined in Sec.~\ref{sec_ipg}, T-2-3D encompasses a two-step conversion process: initial Text-to-Image (T-2-I) generation followed by Image-to-3D (I-2-3D) generation. To improve 3D model creation quality, we employ a background removal module~\footnote{https://github.com/danielgatis/rembg} on T-2-I outputs before I-2-3D processing. The T-2-3D function is:

\begin{equation} 
T23D(P_i) = I23D \circ \text{rembg} \circ T2I(P_i) 
\end{equation}

In detail:
\textbf{(1) Text-to-Image model $T2I(*)$}: Each prompt $P_i$ generates a 2D image $G_{i}=T2I(P_i)$.
\textbf{(2) Background Removal $\rm rembg(*)$}: The generated image $G_i$ undergoes background removal $G'_i={\rm rembg}(G_i)$, enhancing the focus on foreground (i.e., the primary subject).
\textbf{(3) Image-to-3D model $I23D(*)$}: The refined image $G'_i$ is input into I-2-3D, producing the 3D model $D_i=I23D(G'_i)$.
This methodology from text prompt to 3D model via an intermediary image phase, particularly with background removal, ensures a more accurate and intent-aligned 3D reconstruction, elevating the overall quality of the generated models. These functions are tools invoked by \textbf{ LMM agent 1}.

\subsection{Draft 3D Model Selection}
\label{sec_3d_Selection}

Then the \textbf{LMM agent 5 for model selection} in \emph{Idea23D} selects the superior draft 3D model $D_{\rm best}$ from the generated set $\{D_0,D_1,...,D_{N-1}\}$ based on the fidelity and relevance to the user's IDEA. This critical step filters out subpar models, ensuring high-quality iterative generations.

\begin{equation}
D_{\text{best}} = f_{\text{select}}(D_i, X', p_{\text{select}})
\end{equation}

Here, $p_{\rm select}$ is the prompt for the Large Multimodal Model (LMM), guiding the selection of the best draft 3D model. 
$f_{select}$ uses specific few-shot prompts for the LMM. It renders six views of each 3D model, combines them into a single image, and then inputs this image into the LMM. The LMM then selects the 3D model that most closely matches the user’s IDEA input to serve as the draft model for the current iteration.
This mechanism compensates for the discrepancy observed when high-quality T-2-I outputs do not necessarily translate into satisfactory I-2-3D models. By assessing the semantic coherence and visual quality of N similar draft models, the LMM identifies the best one. This comparative analysis, akin to a \emph{find the difference} task, is impossible to achieve by conventional techniques, yet state-of-the-art LMMs like GPT-4V~\cite{openai2023gpt4v}, GPT-4o~\cite{openai_hello_gpt4o} and InternVL~\cite{chen2024expanding-InternVL25} have demonstrated reliable performance in this selection process.

\subsection{Feedback Generation}
\label{sec_Feedback}

After identifying the best draft model $D_{\rm best}$, the \textbf{LMM agent 5} decides on whether to finalize this model as the result $D^*$ or proceed with refinement. In the latter case, the goal is to generate textual feedback $F$ to guide enhancements for $D_{\rm best}$. This decision hinges on whether the iteration count exceeds a maximum threshold $T$ or if the agent believes no further modifications are needed. 

\begin{equation}
F = f_{\text{LMM}}(D_{\text{best}}, X', m, p_{\text{fb}})
\end{equation}

Here, $p_{\rm fb}$ is the LMM prompt for feedback generation, and $m$ denotes the Memory module (discussed in Sec.~\ref{sec_Memory_Module}). This \textbf{LMM agent 6 for feedback generation} assesses discrepancies between $D_{\rm best}$ and the user IDEA $X'$, summarizing key inconsistencies. $D_{\rm best}$ is converted into multi-view images using the $\rm CM2I(*)$ function (from Sec.~\ref{sec_ipg}), aiding the LMM in pinpointing and suggesting specific enhancements. This step is crucial for refining the 3D model. Our experience suggests that clearly defining the aspects for review in $p_{\rm fb}$ significantly enhances the quality of the resultant 3D model.

\subsection{Revised Prompt Generation}
\label{sec_Revised}

In the final stage of each iteration (noted as $\rm iter$), the \textbf{LMM agent 1} comes to the stage again for \emph{Revised Prompt Generation}. It uses textual feedback $F$ and the memory module $m$ to create $N$ refined 3D model generation prompts $\{P_0^{\rm iter+1},P_1^{\rm iter+1}, ... ,P_{n-1}^{\rm iter+1}\}$. This step aims to enhance 3D models generated in the next iteration.

\begin{equation}
P_i^{\rm iter+1} = f_{\text{LMM}}(F, m, X', p_{\text{gen}})
\end{equation}

Here, $p_{\rm gen}$ is the LMM prompt for I-2-3D prompt generation, which is the same with Eq.~\ref{lmm1}, despite the inputs are augmented with $F$ and $m$. Note an LMM agent can readily handle different inputs by prompting it that there are two different cases: the initialization case without feedback and memory and the refinement case with feeback and memory. Agent 1 leverages the information stored in $m$ and the previous iteration's feedback $F$ to generate improved prompts that effectively address the issues identified in $F$. For instance, if feedback $F$ indicates specific visual inaccuracies in the best-to-date model, the revised prompts will focus on rectifying these details through enhanced descriptions. 

\begin{figure*}[t]
  \centering
    \includegraphics[width=0.93\linewidth]{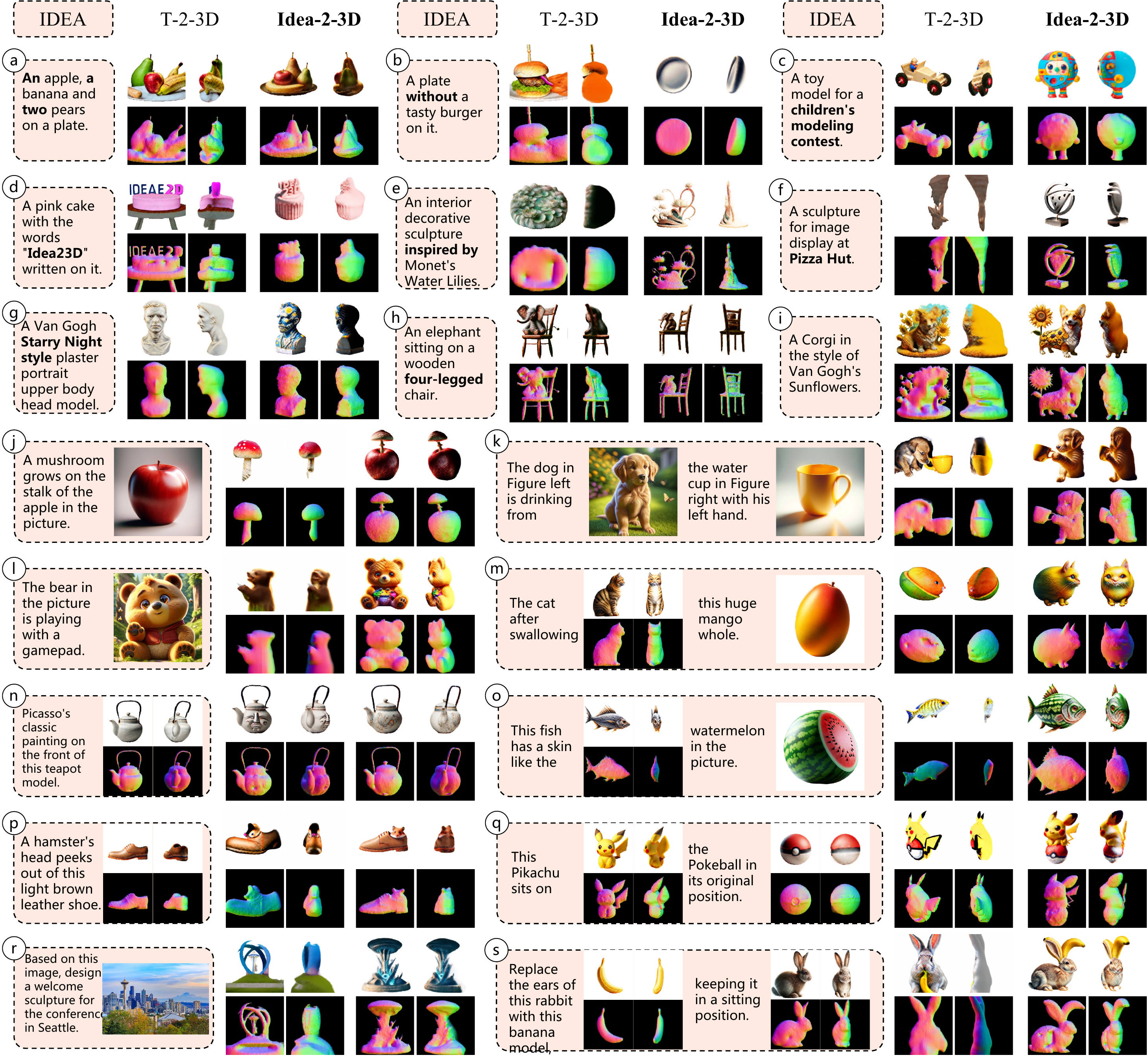}
    \caption{
    Overview of 3D models generated from various types of multimodal IDEA inputs supported by \emph{Idea23D}. 
    The light red box on the left is the user input IDEA containing text, images and 3D models. 
    In the center are the baseline results generated directly from the same T-2-I model with caption-based T-2-I prompt (see Sec.~\ref{Experimental_Setup}). 
    The model on the right is the result generated by iteratively self-refining the T-2-I prompts with \emph{Idea23D}. 
    Comparison with more existing methods is shown in Fig.~\ref{fig:c_models}.
    }
    \label{fig:sample_demos}
\end{figure*}

\subsection{Memory Module}
\label{sec_Memory_Module}

The memory module $m$ is integral to the \emph{Idea23D} framework, serving as a repository for data accrued over the iterative process, as shown by the blue rectangle in Fig.~\ref{fig:overview_famework}. It stores feedback, selected draft 3D models, and corresponding text prompts in a structured 3D model-image-text sequence, enabling the LMM to leverage past experiences and insights gained during previous iterations.

\begin{equation}
m_t = \{P_0^*, D_0^*, F_0, \ldots, P_{t-1}^*, D_{t-1}^*, F_{t-1}\}
\end{equation}

In this representation, $P_{\rm iter}^*,D_{\rm iter}^*,F_{\rm iter}$ represent the optimal text prompt, 3D model, and textual feedback from each $\rm iter^{th}$ iteration. The memory module $m$ aids the agent in identifying specific T-2-3D output traits, such as misunderstood keywords. This knowledge is then integrated into generating refined 3D model prompts, enhancing the precision and adaptability of model generation. If T-2-3D struggles with certain design aspects, $m$ guides subsequent iterations to optimize prompts more effectively, ensuring continuous improvement in \emph{Idea23D}, thereby increasing its alignment with complex IDEAs.

\begin{table*}[t]
    \caption{
We conducted experiments on the \textbf{Eval3DAIGC-198} dataset with the configuration of generating one image per prompt ($num_{img} = 1$), three prompts per round ($num_{draft} = 3$), and up to five iteration rounds ($max_{iters} = 5$). We used GPT-4o~\cite{openai_hello_gpt4o} instead of GPT-4-Vision-Preview~\cite{openai2023gpt4v} from earlier studies. The models used include FLUX.1-dev~\cite{blackforestlabs2024flux1tools}, SD-XL 1.0 with refinement, Hunyuan3D-1.0~\cite{yang2024tencenthunyuan3d}, InternVL2.5-78B~\cite{chen2024expanding-InternVL25}, and LLaVA-CoT-11B~\cite{xu2024llavacot}.
    }
    \label{tab:result_tab}
    \centering
    \resizebox{1.6\columnwidth}{!}{
        \begin{tabular}{ccc|c|cc|cc}
        \toprule
        \multicolumn{3}{c|}{Comparison stage: T-2-3D vs.} & \textbf{Avg. Iter.} & \multicolumn{2}{c|}{CLIP $\uparrow$} & \multicolumn{2}{c}{ULIP-2 $\uparrow$}  \\
        \hline
        \textbf{LMM} & \textbf{T-2-I} & \textbf{I-2-3D} & & T-2-3D & Idea23D & T-2-3D & Idea23D \\
        \hline
        
        \rowcolor{gray!20} GT prompt & FLUX & InstantMesh & - & 0.3152 & - & 0.3134 & - \\
        GPT-4o & FLUX & InstantMesh                       & 1.49 & 0.3003 & 0.3078 &  0.2733 & 0.2917 \\
        InternVL2.5 & FLUX & InstantMesh                  & 2.56 & 0.2896 & 0.3035 & 0.2379 & 0.2756 \\
        LLaVA-CoT & FLUX & InstantMesh                    & 4.67 & 0.2783 &  0.2826 &  0.2108 & 0.2125 \\
        \rowcolor{gray!10} Text Only & FLUX & InstantMesh & - & 0.2745 & - & 0.2056 & - \\
        \hline
        
        \rowcolor{gray!20} GT prompt & FLUX & Hunyuan3D & - & 0.3085 & - & 0.3057 & - \\
        GPT-4o & FLUX & Hunyuan3D                       & 1.41 & 0.2943 & 0.3010 & 0.2675 & 0.2869 \\
        InternVL2.5 & FLUX & Hunyuan3D                  & 2.68 & 0.2870 & 0.2970 & 0.2534 & 0.2783 \\
        LLaVA-CoT & FLUX & Hunyuan3D                    & 4.48 & 0.2734 & 0.2768 & 0.2189 & 0.2273 \\
        \rowcolor{gray!10} Text Only & FLUX & Hunyuan3D & - & 0.2700 & - & 0.2115 & - \\
        \hline
        
        \rowcolor{gray!20} GT prompt & SDXL & InstantMesh & - & 0.2972 & - & 0.2684 & - \\
        GPT-4o & SDXL & InstantMesh                     & 2.02 & 0.2845 & 0.3001 & 0.2302 & 0.2599 \\
        InternVL2.5 & SDXL & InstantMesh                & 3.49 & 0.2810 & 0.2822 & 0.2196 & 0.2211 \\
        LLaVA-CoT & SDXL & InstantMesh                   & 4.53 &  0.2707 & 0.2726 & 0.1961 & 0.1991 \\
        \rowcolor{gray!10} Text Only & SDXL & InstantMesh & - & 0.2680 & - & 0.1979 & - \\
        \hline
        
        \rowcolor{gray!20} GT prompt & SDXL & Hunyuan3D & - & 0.2941 & - &  0.2479 & - \\
        GPT-4o & SDXL & Hunyuan3D                       & 2.14 & 0.2782 & 0.2903 & 0.2118 & 0.2404 \\
        InternVL2.5 & SDXL & Hunyuan3D                  & 3.94 & 0.2725 & 0.2828 & 0.2143 & 0.2198 \\
        LLaVA-CoT & SDXL & Hunyuan3D                    & 4.32 & 0.2711 & 0.2735 & 0.1911 & 0.1956 \\
        \rowcolor{gray!10} Text Only & SDXL & Hunyuan3D & - & 0.2663 & - & 0.1924 & - \\
        
        \bottomrule
        \end{tabular}
    }
\end{table*}

\section{Experiments}
\subsection{Experimental Setup}\label{ExperimentalSetup}
\label{Experimental_Setup}
In our early experiments, the LMMs used were GPT-4V~\cite{openai2023gpt4v} and LLaVA~\cite{liu2023improved}. Fig.~\ref{fig:sample_demos}, Fig.~\ref{fig:c_models}, Fig.~\ref{fig:round_experiment}, Fig.~\ref{fig:democase_detail}, and Fig.~\ref{fig:domain} use GPT-4V as the LMM, DALLE~\cite{openai2023dalle3} as the T-2-I model, and zero123~\cite{liu2023zero} as the I-2-3D model. The results of the User Study, shown in Tab.~\ref{tab:result_tab_userstudy}, were obtained from these early experiments. With the emergence of new LMM, T-2-I, and I-2-3D models, we have also tested Idea23D with these new methods. Fig.~\ref{fig:example} and Fig.~\ref{fig:morecase} use GPT-4o as the LMM, FLUX as the T-2-I model, and InstantMesh as the I-2-3D model. The quantitative results of the ablation study, presented in Tab.~\ref{tab:ablation}, were also based on these new models. This further demonstrates that Idea23D exhibits excellent compatibility with the development of LMMs, T-2-I, and I-2-3D models.

\textbf{T-2-3D baseline.} Our first baseline is caption-based. Since \textbf{no former} 3D AIGC methods can be used for IDEA input, we convert image inputs and 3D model inputs (after multiview rendering) into textual descriptions by captioning them with the LMM. 
In Fig.~\ref{fig:sample_demos}, Tab.~\ref{tab:result_tab} and Tab.~\ref{tab:result_tab_userstudy}, this baseline is called T-2-3D. 
All T-2-3D baselines only use LMM for generating captions and do not perform iterative refinement.

\textbf{Ours w/o iterative refinement.} To demonstrate the impact of an iterative self-refinement design, we construct an \emph{Idea23D} variant with only one iteration. Compared with caption-based baselines, this one features multiple prompt generation and best 3D model selection. 
We presented the results of this part in our early user study, see Sec.~\ref{user_study}.

\subsection{Evaluation Dataset}\label{EvaluationProtocol}
In our experiments, we found that there is a lack of methods to align Text, Image, and 3D in the evaluation of LMM's 3D AIGC capabilities~\cite{liu2023bolaa,andriushchenko2024agentharm,guo2024iwbench,wu2023smartplay,zhujiu2023}. Therefore, following the evaluation practices of Parti~\cite{Yu2022ScalingAM}, we constructed a dataset for evaluating 3D AIGC tasks, called \textbf{Eval3DAIGC-198}, which involves 198 different multimodal interleaved inputs of IDEA, including examples shown in Fig.~\ref{fig:example}, Fig.~\ref{fig:sample_demos} and Fig.~\ref{fig:morecase}. The distribution and examples of the dataset can be found in Appendix~\ref{append:evaluation_dataset}. The cases in the evaluation dataset consist of a text prompt, which may include images or 3D models. Since the results of 3D AIGC are difficult to represent by a specific 3D model as a standard answer, we annotated a description of the 3D model to be generated for each case based on the textual instructions, which serves as the Ground Truth.

\subsection{3D-Caption Quantitative Results}
To ensure a fair and comprehensive comparison of the alignment between user inputs and the final generated 3D models, we employed various methods to convert the generated 3D models into textual descriptions. These textual features were then compared with the handwritten 3D annotations in the Eval3DAIGC-198 evaluation dataset.

(1) CLIP~\cite{radford2021learning} Metric. We render the 3D models from four views (front, back, left, right) and calculate the CLIP similarity between the text description and the rendered images. The CLIP similarity is:

\begin{equation}
S_{\text{CLIP}}(T, I) = \frac{1}{4} \sum_{i=1}^{4} \frac{E_T \cdot E_{I_i}}{\|E_T\| \|E_{I_i}\|}
\end{equation}

where $E_T$ and $E_{I_i}$ are the embeddings extracted from the CLIP model for the text description $T$ and the rendered images $I_i$, respectively. Here, \( \cdot \) denotes the dot product and \( \| \cdot \| \) is the L2 norm.

(2) ULIP-2~\cite{xue2024ulip} Metric. ULIP-2 is a tri-modal pre-training framework that generates text descriptions for 3D shapes without human annotations. It evaluates the alignment between 3D models and texts by comparing 3D shape features with corresponding descriptions in the Eval3DAIGC-198 dataset.

The quantitative experimental results are shown in Tab.~\ref{tab:result_tab}. The "GT prompt" row represents the scores from generating 3D models using manually annotated 3D captions with text-to-image and image-to-3D methods. "Text Only" refers to using only textual instructions from the dataset as the baseline. The "T-2-3D" column shows results where LMM generates descriptions of images and 3D models, which are then combined with dataset instructions to generate 3D models. "Idea23D" represents our proposed method, and "Avg. Iter." shows the average number of optimization iterations for Idea23D.

The results demonstrate that Idea23D improves success rate and accuracy in 3D AIGC tasks, producing outputs closer to ground truth (GT prompt) with fewer iterations. Using GPT-4o as LMM, only two iterations are needed to generate realistic 3D models.
Fig.~\ref{fig:c_models} compares Idea23D with current commercial models, showing that Idea23D achieves new 3D AIGC capabilities not possible with existing models.

\subsection{User Study}\label{user_study}
We presented the results of the caption-based T-2-3D baseline and \emph{Idea23D} to the participants in our user study. Our evaluation results are detailed in Tab.~\ref{tab:result_tab_userstudy}, which compares the caption-based T-2-3D baseline with \emph{Idea23D} using the model specified in each row. We asked users to assess which model (T-2-3D model, first-round results of \emph{Idea23D}, and final-round results of \emph{Idea23D}) was more satisfactory, and to evaluate whether each 3D model complied with the IDEA. Detailed User Study results and explanations can be found in Appendix~\ref{append_userstudy}.

\subsection{Visualization of Self-iterative Refinement}
Fig.~\ref{fig:round_experiment} showcase the evolution and refinement of 3D models at different \emph{Idea23D} stages and how the caption-based baseline works.
Fig.~\ref{fig:democase_detail} shows the iterative self-improvement process of a case in \emph{Idea23D}.
These visual representations illustrate the framework's effectiveness in aligning with user IDEAs and the incremental improvements achieved through its iterative process.

\subsection{Ablation Study}\label{Ablation_study}

\begin{figure}[htbp]
    \centering
    \small
    \includegraphics[width=\columnwidth]{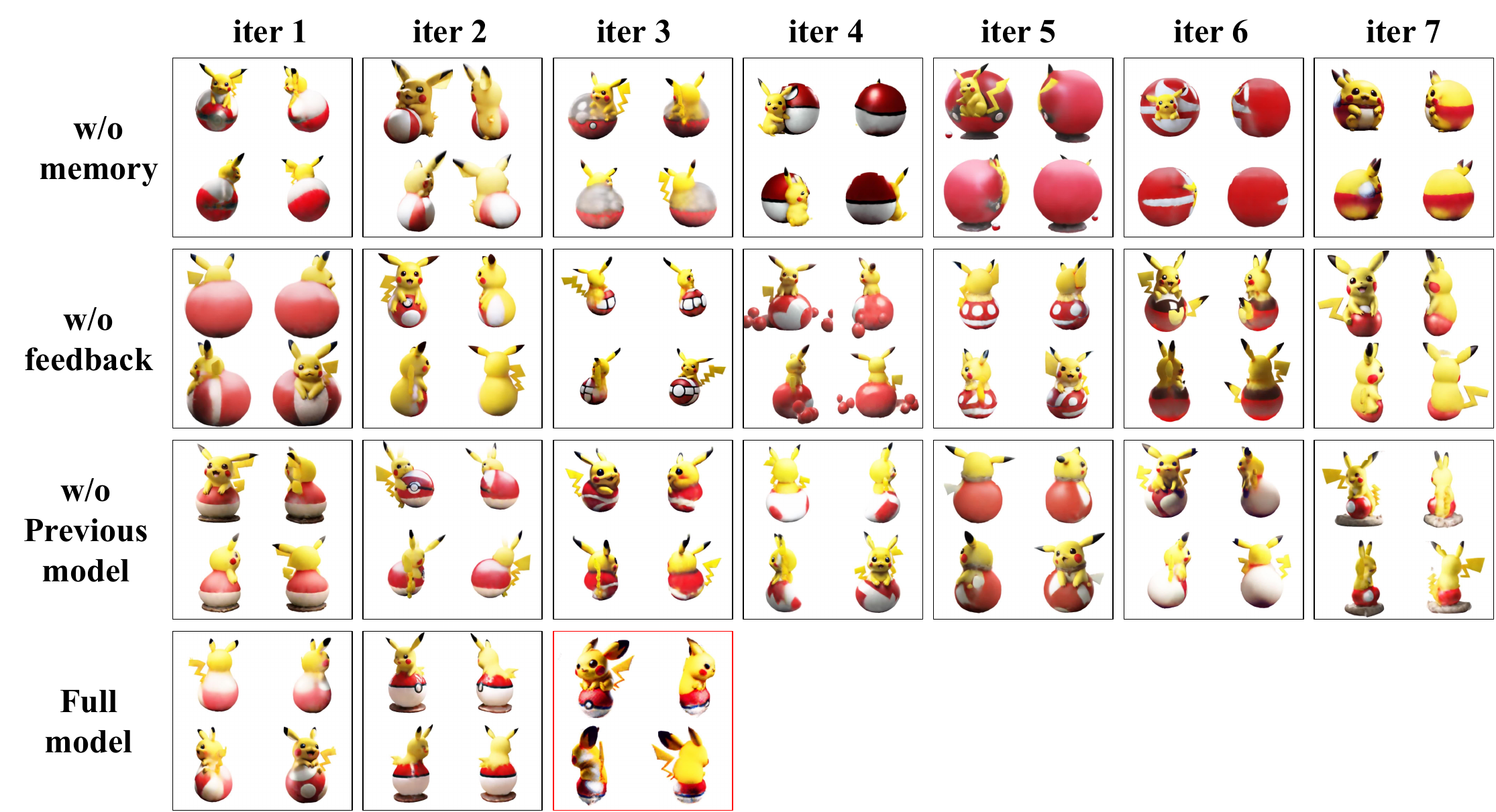}
    \caption{
    Key module ablation across iterations. Note that the full model reaches satisfactory results (judged by the LMM agent 5 in Fig.~\ref{fig:overview_famework}) at iteration 3. The illustrated experiment has no maximum iteration limit.
    }
    \label{fig:ablation}
\end{figure}

We conducted ablation experiments using the first 38 cases from the dataset, with LMM using GPT-4o, T2I using FLUX, and I23D using InstantMesh Iteration rounds and configurations are consistent with Tab.~\ref{tab:result_tab}. Results are shown in Tab.~\ref{tab:ablation}.
The case in Fig.~\ref{fig:ablation} uses GPT-4V, SD-XL, and zero123.

As shown in the Pokemon case (Fig.~\ref{fig:ablation}), removing key modules slows convergence, while the full \emph{Idea23D} model generates a satisfactory result within 3 iterations. The quantitative study in Tab.~\ref{tab:ablation} reveals: \textbf{(1)} The memory module prevents quality divergence, \textbf{(2)} The LMM feedback agent accelerates convergence, \textbf{(3)} Removing stored information from previous models lowers the quality limit after convergence.
Additionally, ablation results show that removing the memory module leads to increasing deviation from user inputs during iterations. Feedback improves convergence speed, and retaining prior models accelerates the process. The ULIP score for 3D models generated with real prompts is 0.3213, with Idea23D reaching 0.3208 after five iterations, while a standard text-to-image pipeline achieves only 0.2830.

\begin{table}[htbp]
\centering
\caption{Ablation study results with the same configuration as in Tab.~\ref{tab:result_tab}.}
\resizebox{1\columnwidth}{!}{
\begin{tabular}{l|c|ccccc}
\toprule
\textbf{} & \multicolumn{1}{c|}{\textbf{Avg. Iter.}} & \multicolumn{5}{c}{\textbf{ULIP$\uparrow$}} \\ 
\cline{3-7}
\textbf{} & \textbf{} & \textbf{Iter 1} & \textbf{Iter 2} & \textbf{Iter 3} & \textbf{Iter 4} & \textbf{Iter 5} \\ 
\hline
w/o memory            & \multicolumn{1}{c|}{1.58} & 0.2953 & 0.2216 & 0.1902 & 0.2139 & 0.1960  \\ 
w/o feedback          & \multicolumn{1}{c|}{1.87} & 0.2930 & 0.2780 & 0.2759 & 0.2773 & 0.2811  \\ 
w/o prev. model       & \multicolumn{1}{c|}{1.78} & 0.2977 & 0.3031 & 0.3066 & 0.3012 & 0.2974  \\ 
\hline
\textbf{full model}            & \multicolumn{1}{c|}{1.49} & 0.3021 & 0.3056 & 0.3108 & 0.3146 & \textbf{0.3208}  \\ 
\hline
\textbf{T-2-3D}            & \multicolumn{1}{c|}{-} & 0.2830 & - & - & - & -  \\ 
\rowcolor{gray!20} GT prompt & \multicolumn{1}{c|}{-} & \textbf{0.3213} & - & - & - & -  \\ 
\rowcolor{gray!10} Text Only  & \multicolumn{1}{c|}{-} & 0.2717 & - & - & - & -  \\ 
\bottomrule
\end{tabular}
}
\label{tab:ablation}
\end{table}

\section{Conclusion}
\emph{Idea23D}, utilizing an LMM agent collaboration framework, revolutionizes 3D AIGC by automating the creation of models from high-level, interleaved multimodal user inputs (IDEAs). This innovative system excels in integrating text, images, and 3D models, underpinned by a unique iterative process that enhances model coherence and visual alignment with IDEAs. User studies underscore its superiority in satisfaction and comparative quality, marking \emph{Idea23D} as a significant advancement in 3D AIGC and a benchmark for future design tools.

\clearpage
\section{Limitations}\label{Limitations}
\emph{Idea23D} effectively improves the alignment between 3D generation models and user intent, but it still relies on LMM and I-2-3D models. As shown in Tab.~\ref{tab:result_tab}, LMM significantly impacts the 3D generation results. However, commercial models like GPT-4V can already generate image prompts based on feedback very well. On the other hand, open-source LMMs like LLaVA still have significant shortcomings in image understanding capabilities.

According to our experiments, the main bottleneck of \emph{Idea23D} at this stage lies in the Image-to-3D step. Even the most advanced Image-to-3D models can fail. In the worst-case scenario with a very low probability, the final model output when \emph{Idea23D} reaches the maximum iteration may still not meet user requirements. Nonetheless, \emph{Idea23D} can ensure that the final selected model is the most aligned with user input among all generated 3D models throughout the iterations.

\section{Acknowledge}
This work was supported by the Fundamental Research Funds for the Central Universities under Grant Number KG16336301 and the China Postdoctoral Science Foundation under Grant Number 2024M764093.

\bibliography{main}

\newpage
\appendix

\section{Appendix: Evaluation Dataset}\label{append:evaluation_dataset}
These IDEAs span a range of complexities: 9 were text-only, 57 featured text and image inputs, 68 included text and 3D model inputs, and 64 contained text, image, and 3D model inputs. Each test case was meticulously designed to represent real-world scenarios. The dataset cases are the same as the IDEAs in Fig.~\ref{fig:example} , Fig.~\ref{fig:sample_demos} and Fig.~\ref{fig:morecase}.

\begin{figure}[htbp]
    \centering
    \includegraphics[width=\linewidth]{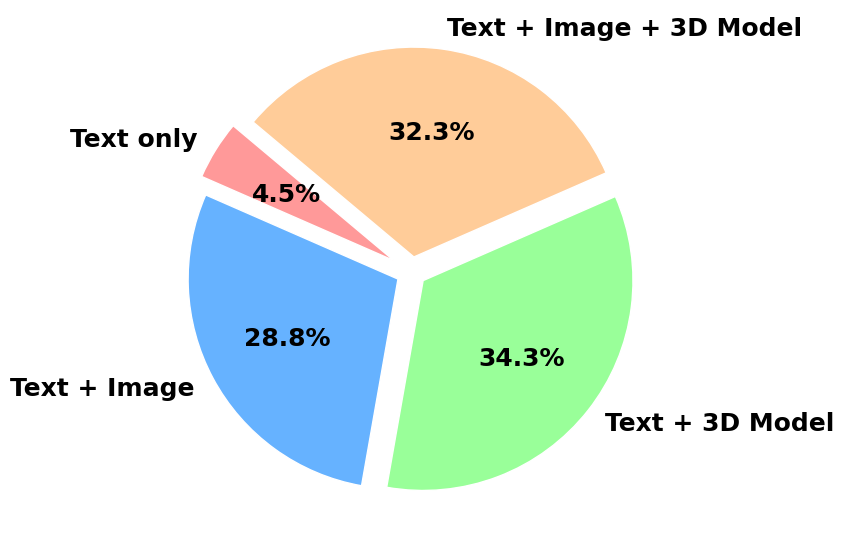}
    \caption{IDEA Content Distribution.}
    \label{fig:idea-content-distribution}
\end{figure}

The dataset also includes a distribution of tags: 9 IDEAs contained 0 tags, 62 IDEAs had 1 tag, and 127 IDEAs included 2 tags, highlighting the diversity of annotation complexities.

\begin{figure}[htbp]
    \centering
    \includegraphics[width=\linewidth]{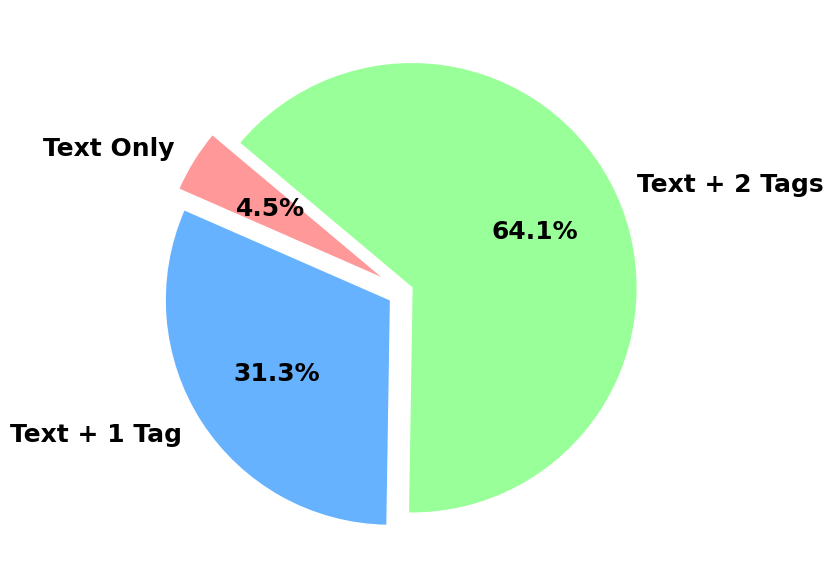}
    \caption{Tag Distribution in IDEAs.}
    \label{fig:tag-distribution}
\end{figure}

\section{Appendix: Visualization Results}

\begin{figure*}[htbp]
    \centering
    \small
    \includegraphics[width=\textwidth]{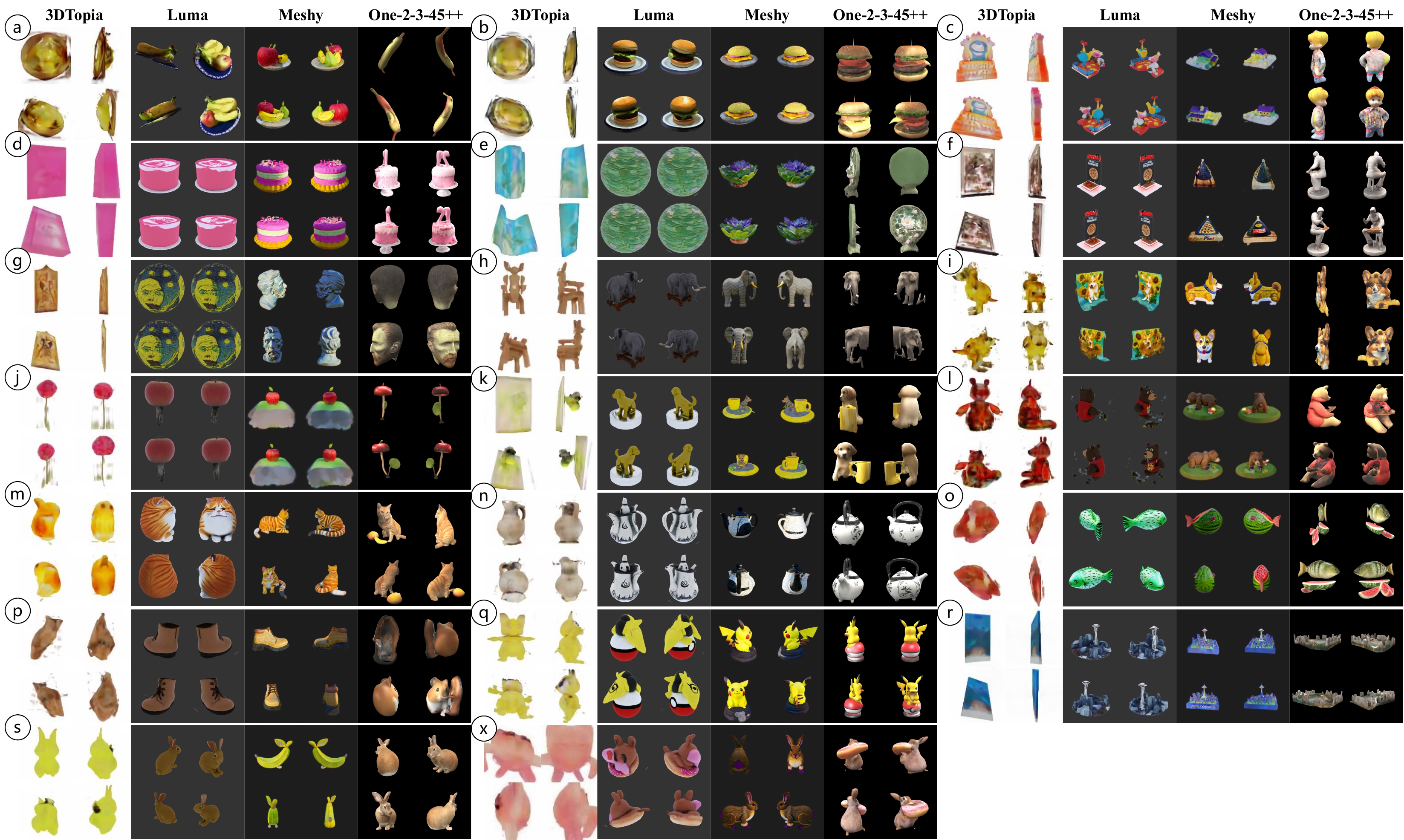}
    \caption{
    Comparison with commercial T-2-3D models. 
    Inputs are the same as Fig.~\ref{fig:sample_demos}.  
    Case (\textit{x}) corresponds to the case of Fig.~\ref{fig:example}.
    }
    \label{fig:c_models}
\end{figure*}

\begin{figure*}[htbp]
    \centering
    \small
    \includegraphics[width=\textwidth]{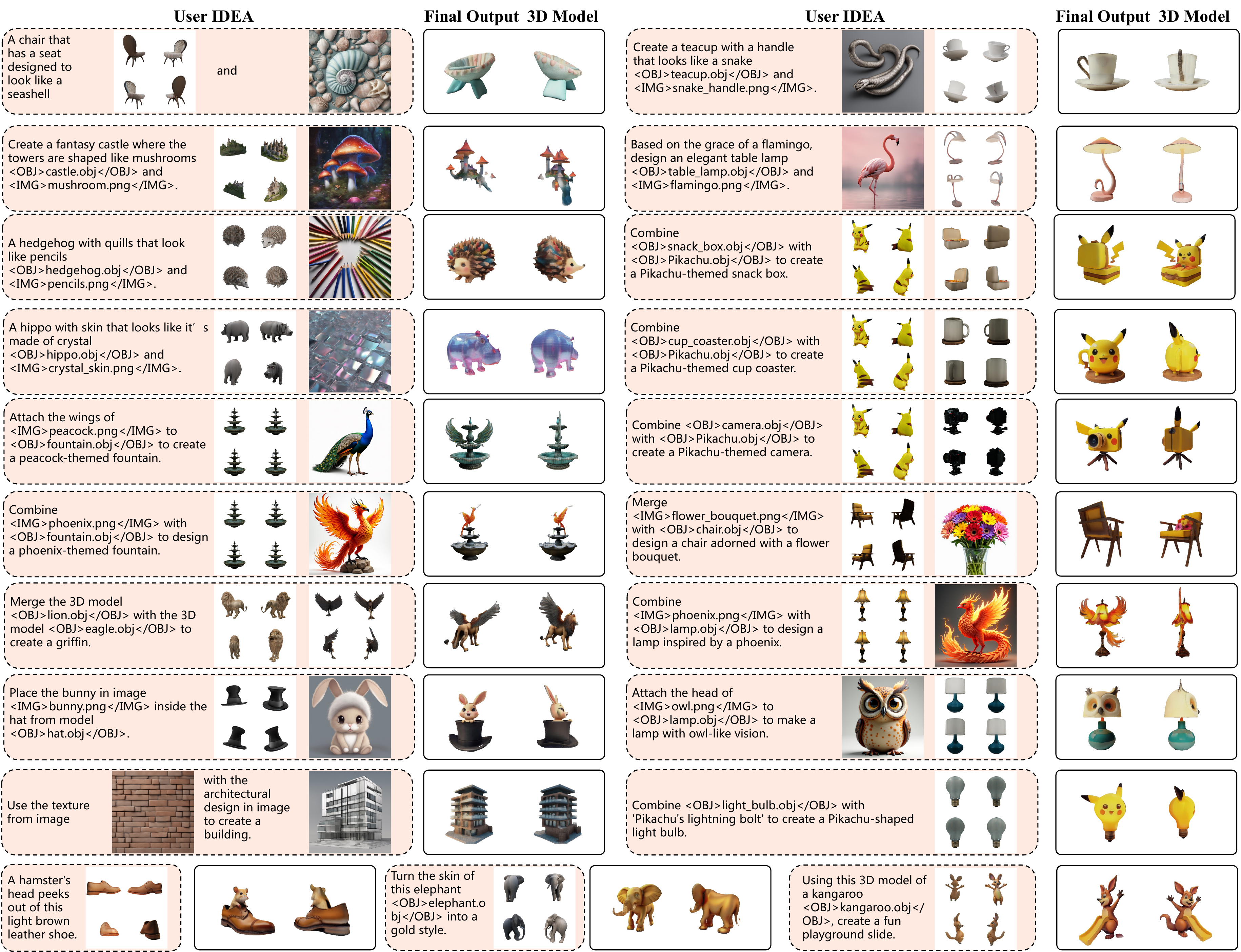}
    \caption{
Results using GPT-4o, FLUX, InstantMesh as Idea23D components, cases using cases from the dataset.
    }
    \label{fig:morecase}
\end{figure*}

The results of the visualization are shown in the Fig.~\ref{fig:example}, Fig.~\ref{fig:c_models} and Fig.~\ref{fig:morecase}.
For Tab.~\ref{tab:result_tab_userstudy}, we evaluate three T-2-I mdoels: DeepFloyd IF~\cite{DeepFloyd2023if}, DALL·E~\cite{openai2023dalle2}, and SD-XL~\cite{podell2023sdxl}, five I-2-3D models: Zero123~\cite{threestudio2023, liu2023zero}, Wonder3D~\cite{long2023wonder3d}, TripoSR~\cite{tochilkin2024triposr}, InstantMesh~\cite{xu2024instantmesh} and LGM~\cite{tang2024lgm}, and two LMM agent options: GPT-4V~\cite{openai2023gpt4, openai2023gpt4} and LLaVA~\cite{liu2023visual, liu2023improved}.

Our framework has \textbf{good compatibility}. After our initial experiments, advanced image-to-3D generation models such as TripoSR~\cite{tochilkin2024triposr}, InstantMesh~\cite{xu2024instantmesh}, and LGM~\cite{tang2024lgm} have emerged.
Tab.~\ref{tab:result_tab} demonstrates that using these state-of-the-art models as the base modules for our T-2-I and I-2-3D results in significant improvements in \emph{Idea23D} generation.

\section{Appendix: User Study Results}\label{append_userstudy}

\begin{table*}[t]
    \caption{
Results of the user study.
    }
    \label{tab:result_tab_userstudy}
    \centering
    \resizebox{2\columnwidth}{!}{
    \begin{tabular}{ccc|cc|cc|cc|cc}
    \toprule
    \multicolumn{3}{c|}{Comparison stage: T-2-3D vs. } & \multicolumn{4}{c|}{\emph{Idea23D} (Initial-round)} & \multicolumn{4}{c}{\emph{Idea23D} (Iterative self-refined)}  \\ \cline{1-11}  
    
    \multicolumn{3}{c|}{\diagbox{Model}
        {\begin{tabular}[c]{@{}c@{}}Evaluation\end{tabular} }} & \multicolumn{2}{c}{Which is better} &\multicolumn{2}{c|}{\begin{tabular}[c]{@{}c@{}}Satisfying  IDEA\end{tabular}}    & \multicolumn{2}{c}{Which is better}  & \multicolumn{2}{c}{\begin{tabular}[c]{@{}c@{}}Satisfying  IDEA\end{tabular}} \\ \cline{1-11}
    
    T-2-I & I-2-3D & LMM&  T-2-3D & \textbf{\emph{Idea23D}}  & T-2-3D   & \textbf{\emph{Idea23D}}  & T-2-3D   & \textbf{\emph{Idea23D}}  & T-2-3D   & \textbf{\emph{Idea23D}}  \\ \cline{1-11} 
    
    SD-XL &      TripoSR &       GPT-4V       &38.5\%   &\textbf{61.5\%}     &  54.7\%    &\textbf{80.2\%}      &18.9\%   &\textbf{81.1\%}    &    -  &\textbf{96.2\%}\\ 
    SD-XL &      InstantMesh &   GPT-4V       &43.2\%   &\textbf{56.8\%}     &  53.1\%    &\textbf{82.9\%}     &20.1\%   &\textbf{79.9\%}    &    -  &\textbf{96.4\%}\\ 
    SD-XL &      LGM &GPT-4V       &41.3\%   &\textbf{58.7\%}     &  56.2\%    &\textbf{81.3\%}     &12.5\%   &\textbf{87.5\%}    &    -  &\textbf{94.5\%}\\ 
    DALL·E &     Zero123 &   GPT-4V    &25.8\%    &\textbf{74.2\%}    &  41.5\%   &\textbf{78.2\%}     &6.5\%  &\textbf{93.5\%}       & -    &\textbf{94.2\%}\\ 
    \hline
    DALL·E &     Zero123 &   LLaVA     &27.5\%   &\textbf{72.5\%}     &  33.2\%   &\textbf{70.6\%}     &29.7\%  &\textbf{70.3\%}  &  -    &\textbf{82.3\%}\\ 
    DALL·E &     Wonder3D  & GPT-4V    &18.8\%    &\textbf{81.2\%}    &  47.9\%    &\textbf{74.3\%}     &3.5\%  &\textbf{96.5\%}   & -   &\textbf{91.1\%}\\
    DALL·E &     Wonder3D  &  LLaVA    &16.3\%   &\textbf{83.7\%}     &  28.3\%    &\textbf{69.4\%}     &10.3\%  &\textbf{89.7\%}      &  -         &\textbf{76.2\%}\\
    DeepFloyd IF &          Zero123 &   GPT-4V    &20.0\%    &\textbf{80.0\%}    &  38.5\%     &\textbf{64.1\%}     &28.3\%  &\textbf{71.7\%}     & -   &\textbf{73.7\%}\\
    DeepFloyd IF &          Zero123 &   LLaVA     &29.0\%    &\textbf{71.0\%}     &  23.1\%    &\textbf{57.6\%}     &34.0\%  &\textbf{66.0\%}     &  -      &\textbf{66.8\%}\\
    DeepFloyd IF &          Wonder3D  &  GPT-4V   &29.3\%    &\textbf{70.7\%}     & 32.6\%    &\textbf{66.1\%}     &14.8\%  &\textbf{85.2\%}     & -   &\textbf{75.9\%}\\ 
    DeepFloyd IF &          Wonder3D  &   LLaVA   &26.0\%    &\textbf{74.0\%}     & 19.3\%          &\textbf{55.3\%}     &18.3\%  &\textbf{81.7\%}     &  -       &\textbf{64.5\%}\\ 
    
    \bottomrule
    \end{tabular}
    }
\end{table*}

Tab.~\ref{tab:result_tab_userstudy} shows the results of our user preference study.
Our user study was conducted by distributing an online web-based survey questionnaire, with over 200 users participating.
This user study's comparative analysis reveals the remarkable superiority of the \emph{Idea23D} framework over existing caption-based T-2-3D methods. Due to the lack of multi-round iterations in the T-2-3D baseline, the empty parts in the "Iterative self-refined" column of the table are the same as those in the "Initial-round" column.

Participants were presented with various IDEAs alongside multiple 3D model outputs from both caption-based T-2-3D baselines and \emph{Idea23D}. 
To aid decision-making, users viewed rotating video representations of each 3D model. The order of presentation for both the cases and the model outputs was randomized to avoid any sequence bias, ensuring an unbiased assessment of the users' preferences.

Our evaluation, detailed in Tab.~\ref{tab:result_tab_userstudy}, compares caption-based T-2-3D baselines and \emph{Idea23D} using models specified in each row. 
We ask users to evaluate which model (the T-2-3D model, the first round results of \emph{Idea23D}, and the end round results of \emph{Idea23D}) is more satifying, as well as evaluating each 3D model for compliance with IDEA.

The results show that \emph{Idea23D} markedly enhances user preference scores across a diverse range of T-2-I, I-2-3D, and LMM models. Notably, the \emph{Idea23D} framework's initial prompting stage significantly outperforms the caption-based T-2-3D results by effectively decomposing and interpreting the user's multimodal IDEA, thereby selecting the most suitable 3D model. This improvement is further amplified in the iterative self-refinement stage of \emph{Idea23D}.

For instance, in scenarios utilizing DALL-E, Zero123, and GPT-4V, \emph{Idea23D} models were preferred by users in 74.2\% of the cases over T-2-3D in the initial round, demonstrating a higher IDEA satisfaction rate (78.2\%). Conversely, T-2-3D models achieved only 41.5\% satisfaction. In the iterative refinement comparisons, \emph{Idea23D} models were favored even more (93.5\% preference), achieving a remarkable 94.2\% satisfaction rate.

These findings were consistent across various T-2-I, I-2-3D, and LMM configurations. Additionally, we observed that stronger T-2-I models, with enhanced language understanding capabilities, contributed to improved performance in \emph{Idea23D}, suggesting that our framework may enjoy the development of off-the-shelf models it invokes.

\section{Appendix: Visualization of Self-iterative Refinement}
\begin{figure*}[t]
  \centering
    \includegraphics[width=0.9\linewidth]{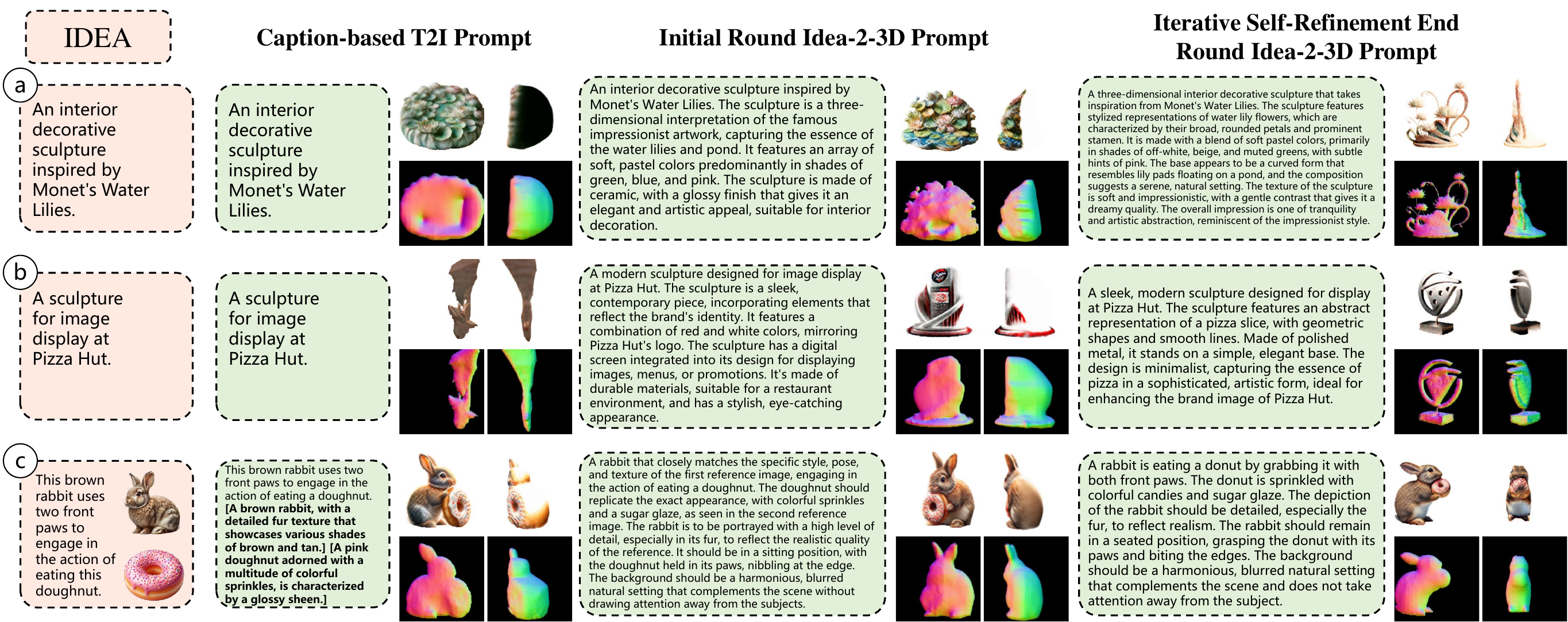}
    \caption{
    Comparison between prompts for "Caption-based T-2-I prompt for initial round", "Idea2-3-D prompt for initial round" and "Iterative self-refinement \emph{Idea23D} end round" and the comparison between generated 3D models.
    }
    \label{fig:round_experiment}
\end{figure*}

\begin{figure*}[t]
  \centering
    \includegraphics[width=0.9\linewidth]{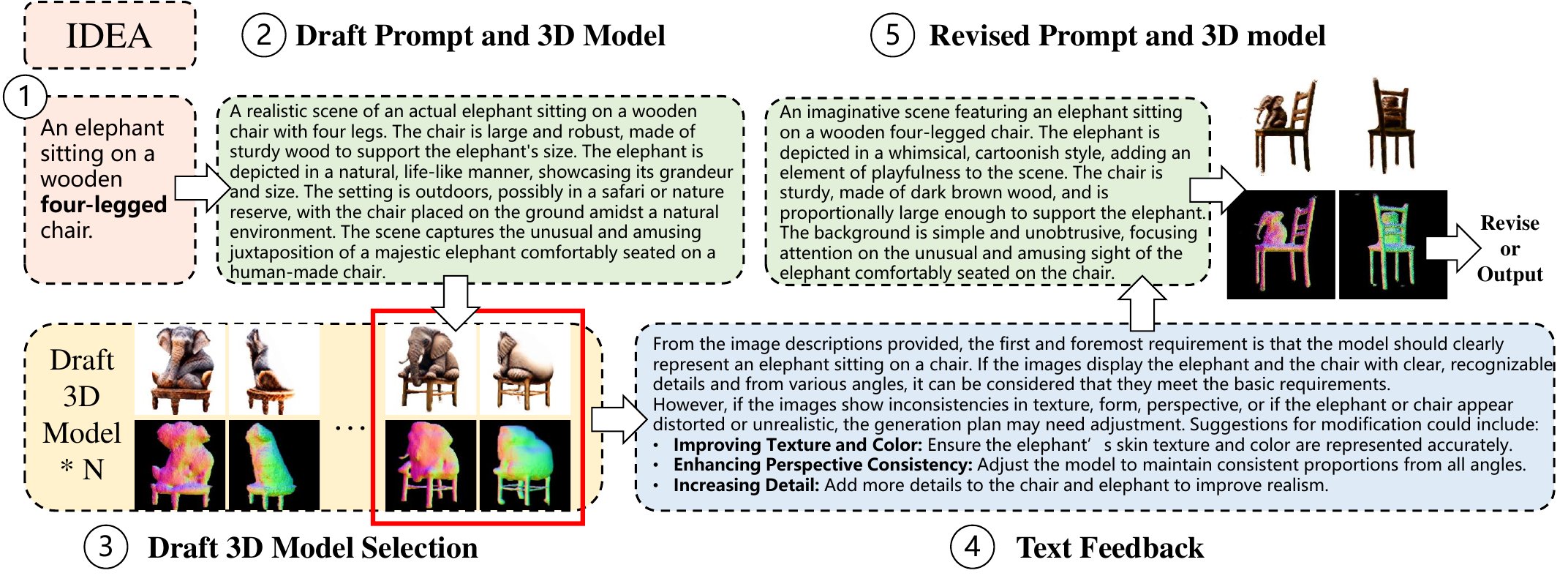}
    \caption{
    An example of the process for a 3D model generated through the \emph{Idea23D} framework. An initial prompt is first generated based on the input IDEA and $p_{\text{gen}}$, then multiple images are generated from the T-2-I model, and then the corresponding draft 3D model is generated for each image separately. The most appropriate 3D model is selected based on $p_{\text{select}}$ and feedback is generated. The results are output after several rounds of iterations. Green rectangles indicate text prompts for T-2-3D. blue rectangles indicate: modification suggestions given by agent based on draft 3D models and $p_{\text{fb}}$.
    }
    \label{fig:democase_detail}
\end{figure*}

Fig.~\ref{fig:round_experiment} showcase the evolution and refinement of 3D models at different \emph{Idea23D} stages and how the caption-based baseline works.
Fig.~\ref{fig:democase_detail} shows the iterative self-improvement process of a case in \emph{Idea23D}.

\section{Appendix: Efficiency} 
Since generation from $N$ prompts can run in parallel, the inference speed is agnostic of $N$. 
\emph{Idea23D} inference speed mainly depends on the speed of the T-2-I and I-2-3D models, as the average number of iterations is 2-3. For example, with the \emph{Idea23D} implemented using GPT-4V + zero123 + DALLE, generating an optimal result takes about 10 minutes (zero123 requires approximately 4 minutes to generate a 3D model from an image).
In the Colab implementation mentioned in our abstract, we use GPT-4V + TripoSR + SDXL to implement the \emph{Idea23D} framework, and it takes about 5 minutes to generate a final 3D model (with 3 iterations).
For reference, existing commercial methods typically generate a model in about 5-10 minutes. For example, 3DTopia~\footnote{https://github.com/3DTopia/3DTopia} and One-2-3-45++\cite{liu2023one} take about 5 minutes on average, while Luma\footnote{https://lumalabs.ai/genie?view=create} and Meshy~\footnote{https://www.meshy.ai/} take about 10 minutes. Recent advanced methods such as TripoSR~\cite{tochilkin2024triposr}, InstantMesh~\cite{xu2024instantmesh}, and LGM~\cite{tang2024lgm} have compressed the time to generate 3D models from images to within 1 minutes.

\section{Appendix: More Specific Domains}\label{moredomains}
\begin{figure*}[htbp]
    \centering
    \small
    \includegraphics[width=\textwidth]{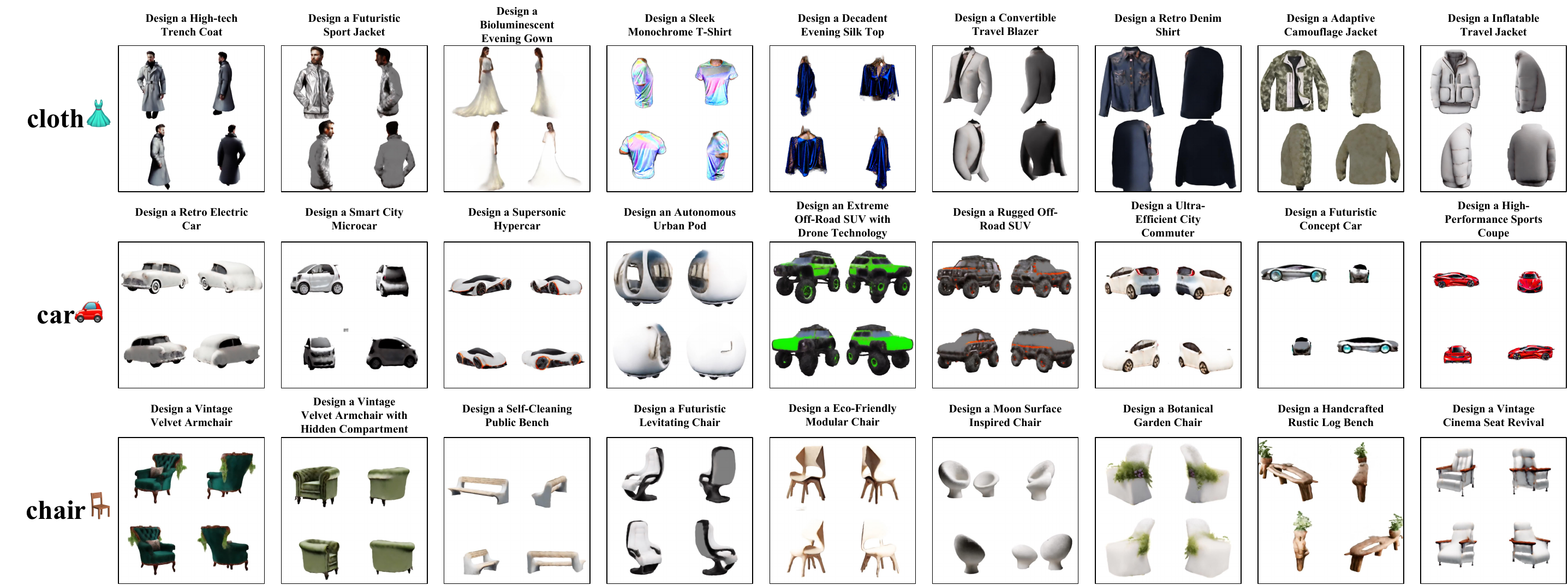}
    \caption{
    Domain-specific results of \emph{Idea23D}.
    }
    \label{fig:domain}
\end{figure*}

We tested our approach within the domain of automated design and modeling of \textbf{car}, \textbf{chair}, and \textbf{cloth}. The four views are plotted as a result of 3D modeling, and the text prompts above indicate the User Idea we entered.
The result is shown in Fig.~\ref{fig:domain}. The abstract IDEA input is above the picture.

\end{document}